\documentclass[a4paper]{article}

\usepackage{INTERSPEECH2019}

\usepackage{amsmath,graphicx,multirow}
\usepackage[noadjust]{cite}
\usepackage{CJKutf8}
\newcommand{\mman}[1]{\begin{CJK}{UTF8}{gbsn} #1\end{CJK}}

\title{On the End-to-End Solution to  Mandarin-English Code-switching\\Speech Recognition}
\name{Zhiping Zeng$^{1}$, Yerbolat Khassanov$^{2}$, Van Tung Pham$^{1}$, Haihua Xu$^{1}$, Eng Siong Chng$^{1,2}$, Haizhou Li$^{3}$}
\address{
  $^1$Temasek Laboratories, Nanyang Technological University, Singapore\\
  $^2$School of Computer Science and Engineering, Nanyang Technological University, Singapore\\
  $^3$Department of Electrical and Computer Engineering, National University of Singapore, Singapore}
\email{zengzp@ntu.edu.sg}

\begin{document}

\maketitle
\begin{abstract}
  Code-switching (CS) refers to a linguistic phenomenon 
where a speaker uses different languages in an utterance or between alternating utterances.
In this work, we  study  end-to-end (E2E) approaches to the 
Mandarin-English code-switching speech recognition task. 
We first examine the effectiveness of using data augmentation and byte-pair encoding (BPE) subword units. 
More importantly, we propose a multitask learning recipe, where a language identification task 
is explicitly learned in addition to the E2E speech recognition task. 
Furthermore, we introduce an efficient word vocabulary expansion method for language modeling to alleviate  data sparsity issues under the code-switching scenario. 
Experimental results on the SEAME data, a Mandarin-English code-switching corpus, demonstrate 
the effectiveness of the proposed methods.
\end{abstract}


\section{Introduction}
\label{sec:intro}
Code-switching (CS) is a linguistic phenomenon, where speaker's utterances contain different languages,
either inside a given utterance or  between utterances. It frequently appears in  world wide areas. 
Therefore, developing a code-switching speech recognition (CSSR) system is important and has received increasing attention recently.

While DNN-HMM-based automatic speech recognition (ASR) framework is popular in code-switching speech recognition  \cite{YilmazHL18,GuoXXC18}, it has some clear limitations. 
Firstly, one needs to build a big lexicon mixed with words from different
languages, and it would take more human efforts to label pronunciations for those words from different languages.
Secondly, acoustic modeling (AM), language modeling (LM) and lexicon modeling components of the  DNN-HMM-based  ASR  system, which  are  optimized  separately.
This would lead to sub-optimal performance.

In this paper, we pursue an End-to-End (E2E) strategy to resolve Mandarin-English code-switching speech recognition instead. In contrast to the DNN-HMM-based approach, it doesn't require any lexicon modeling efforts. More importantly, the entire recognition system
comprises compactly connected neural networks that are jointly learned from scratch. To our best knowledge,
this is the first attempt of using E2E strategy to code-switching speech recognition task. Our contributions mainly lies in the following aspects.

Firstly, we manage to build competitive E2E baseline systems
using data augmentation and byte-pair encoding (BPE) based subword units \cite{DBLP:journals/corr/abs-1805-03294,sennrich2015bpe,kudo2018subword}.
We found data augmentation is more effective to the E2E framework than the DNN-HMM-based one.
Besides, we found the BPE subword units
yield better recognition results than characters. This is consistent with
what is reported in \cite{DBLP:journals/corr/abs-1805-03294}.

Secondly, we employ a multitask learning (MTL) \cite{pan2010survey} method to enhance our E2E-based 
code-switching ASR system. Specifically, we propose to use language identification (LID) as the auxiliary task to help improving the speech recognition performance.
It showed that the LID-based MTL helps on Token Error Rate (TER) reduction. 

Thirdly, 
to alleviate the cross-lingual data sparsity issue in language modeling,
we introduce a word vocabulary expansion method
inspired by \cite{khassanov2018unsupervised}. Note that in \cite{khassanov2018unsupervised}, this technique is applied for monolingual data to
improve the speech recognition performance when its output is rescored by language modeling,
while this work applies it to the code-switching language model rescoring.

The  paper  is  organized  as  follows. Related work are presented in Section \ref{sec:Related_work}. Then,
proposed multitask learning E2E code-switching speech recognition approaches and code-switching word vocabulary expansion LM rescoring are introduced in Section \ref{sec:proposed_approaches}. 
Experimental setups are described in Section \ref{sec:exp}, then experimental results and analysis are reported in Section \ref{sec:exp_results}. Finally, we conclude and talk about future work in Section \ref{sec:conclusion}.

\section{Related work}
\label{sec:Related_work}
Previous works on code-switching speech recognition relied on the conventional GMM-HMM or DNN-HMM framework \cite{YilmazHL18,GuoXXC18,WesthuizenN17,vu:icassp12}. 
Recently, 
end-to-end speech recognition methods have drawn much attention and produced promising recognition results \cite{graves2006connectionist,chan2016listen,GravesJ14,MaasHJN14,HannunCCCDEPSSCN14,DBLP:journals/corr/BahdanauCB14,chiu2017state,watanabe2017hybrid}. 
There are two main directions for end-to-end speech recognition. One is the earlier proposed
Connectionist
Temporal Classification (CTC) \cite{graves2006connectionist},  and the other is the attention mechanism based method \cite{chan2016listen}. 
CTC performed well on various corpora such as WSJ \cite{GravesJ14,MaasHJN14} and SWB \cite{HannunCCCDEPSSCN14}. 
Recently, inspired from the attention-based machine translation framework \cite{DBLP:journals/corr/BahdanauCB14},
attention-based E2E began to play a crucial role in speech recognition, 
achieving the state-of-the-art 
results \cite{chiu2017state}.
Despite the two methods are different, researchers in \cite{watanabe2017hybrid} exploited the advantages of each methods to
build the hybrid CTC/attention based E2E ASR system,
which leads to better results compared to an ASR system built with
either single method.

We apply end-to-end approach to code-switching speech recognition task in this paper. 
This differs from the previous end-to-end speech recognition that has worked only for monolingual case.
Recently, it was shown that E2E method can be employed to 
perform  multilingual speech recognition simultaneously \cite{DBLP:conf/asru/WatanabeHH17,Toshniwal}. However, the multilingual speech recognition is not a code-switching task.

\section {Approaches to end-to-end CSSR}
\label{sec:proposed_approaches}
In this section, we present various approaches to 
achieve better end-to-end code-switching speech recognition.
We first aim to build  competitive baselines.
To start with, we investigate data augmentation method and study different subword units as the output of the end-to-end CSSR system. After that, we
employ a multitask learning by introducing
the LID as the auxiliary task to
boost the performance of our end-to-end CSSR. Finally, we propose
a modified neural language modeling framework, aiming to alleviate
the cross-lingual data sparsity issue within the 
CSSR task.

\subsection{Approaches to develop E2E CSSR baselines}
\label{sssec:subword}
One major challenge of building the E2E system is that it requires a lot of data to train the model \cite{ramachandran2016unsupervised}. To deal with this problem, we apply 
speech speed perturbation based data augmentation method proposed in \cite{Ko2015AudioAF,ko2017aug}, as the effectiveness of the method has been proved in the conventional DNN-HMM ASR method. By manipulation, we 
obtained x3 times of the original data, with a speaking rate of 
90\%, 100\%, and 110\% of the original data respectively. In this work, E2E with data augmentation is one of our baseline. 

Another issue for E2E system is how to select the 
output units. 
As we are dealing with
both Mandarin and English output units simultaneously, it sounds straightforward to use characters as the units.
However, this will result in only 26 output units for English
and several thousand for Mandarin. 
We conjecture that such an unbalanced situation will be disadvantageous to English, yielding worse recognition results.
To balance the units between the two languages, we decide to use subword units \cite{chiu2017state} for English. BPE subword units have shown to be helpful for English \cite{DBLP:journals/corr/abs-1805-03294}. As a result, in this work, we use the BPE subwords for English, while leaving 
the output units of Mandarin fixed with characters. Therefore, another baseline in this work is to apply BPE subword units on the E2E system trained with data augmentation.

\subsection{Multitask Learning of CSSR with LID}
\label{sssec:MTL}

Inspired from the work in \cite{watanabe2017hybrid}, we adopted a hybrid CTC/\hspace{0pt}attention based 
 E2E architecture to conduct CSSR.
 
 Specifically, let $X$ be the input acoustic sequence, $Y$ be the output sequence comprising characters or BPE units, $\mathcal{L}_{CTC}(Y|X)$ be the CTC objective loss, $\mathcal{L}_{att}(Y|X)$ be the attention-based objective loss.   
 Then,
 the  objective loss $\mathcal{L}_{MTL} (Y|X)$ of the entire hybrid E2E system is as follows:
{
\begin{equation}\label{eqn:hybrid-loss}
\mathcal{L}_{MTL} (Y|X)=  \lambda_1\mathcal{L}_{att}(Y|X)
+ (1- \lambda_1)\mathcal{L}_{CTC}(Y|X)
\end{equation}}
where $\lambda_1 \in [0,1]$ is a hyper-parameter to control the contribution of each model. 

\begin{figure}[htbp]
  \centering
  \includegraphics[width=\linewidth]{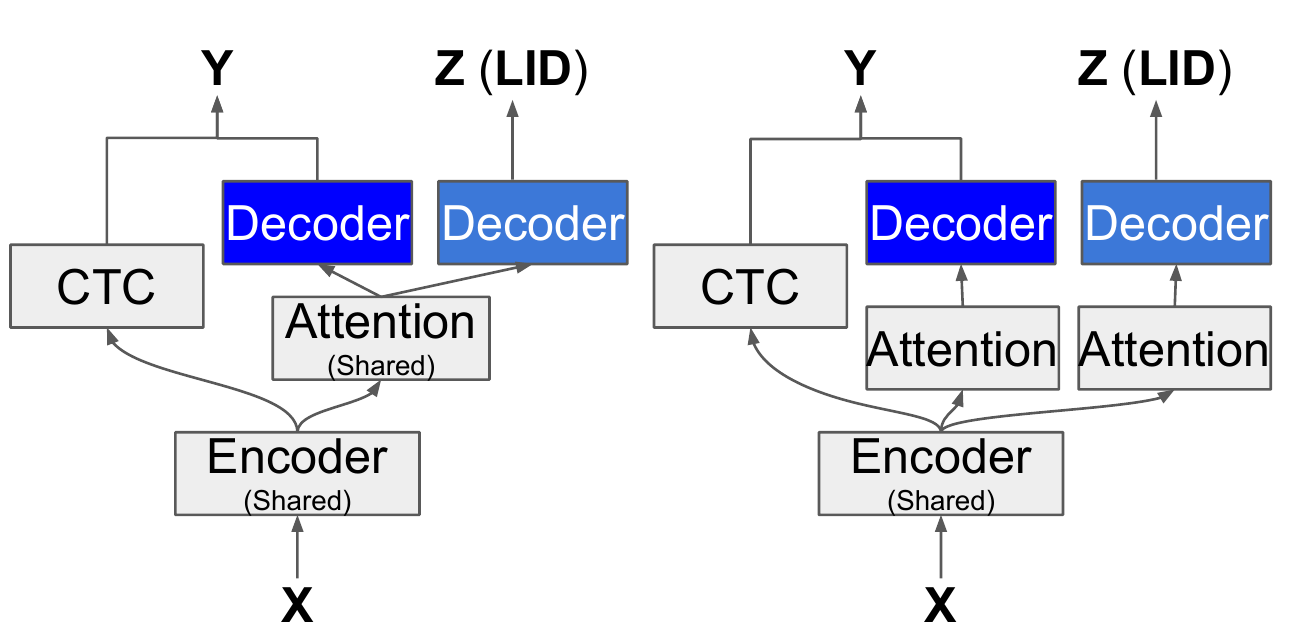}
  \caption{A multitask learning (MTL) framework with language identification (LID) for E2E-based CSSR. The LID task can share the same attention module (left) or use a separate attention module (right).}
  \label{fig:e2e_archi_lid}
\end{figure}

For the code-switching system, although we can infer the language identification from the decoding transcription, the language information has not been used explicitly during training. We believe that using the language identification would improve the CSSR performance. To this end, we extend the multitask learning method as indicated in Eq (\ref{eqn:hybrid-loss}).
The framework is illustrated in Figure \ref{fig:e2e_archi_lid}.
As a result, the whole training objective loss is changed as follows:
{
\begin{equation}\label{eqn:mtl-loss}
\begin{split}
\mathcal{L}_{MTL} (Y|X)=\ & \lambda _{1}\mathcal{L}_{att}(Y|X)+ (1- \lambda _{1})\mathcal{L}_{CTC}(Y|X) \\
 & +\lambda _{2}\mathcal{L}_{lid}(Z|X) 
 \end{split}
\end{equation}}
where $Z$ is the output LID sequence, $\mathcal{L}_{lid}(Z|X)$ represents the LID objective loss, and we restrict $\lambda_2 \in [0,1]$.

As indicated in Figure \ref{fig:e2e_archi_lid}, we have investigated
two methods to incorporate the LID into the hybrid CTC/attention framework.
One is to share the same attention model with the speech recognition task ({\it LID}$_{shared}$),
and another is to learn an independent attention model by itself ({\it LID}$_{indep}$).
Both methods use the same objective function as in the Eq (\ref{eqn:mtl-loss}). 

\subsection{Vocabulary expansion for neural language model}
\label{sssec:LM}
The performance of E2E speech recognition can be further improved 
when its output is rescored by neural language model (NLM)~\cite{chan2016listen}.
However, the vocabulary coverage of the NLM is usually a shortened list of 
the entire ASR vocabulary, and such the list is usually only composed 
of most frequent words, due to the necessity of 
learning complexity restriction. 
Consequently, 
the probability of those infrequent words that are out-of-shortlist 
are not well learned by the NLM.

In code-switching speech recognition scenario, we treat those words that occur at the cross-lingual 
transition positions as `infrequent' words, in addition to 
other `infrequent' words that we have in monolingual text.
The main idea is to `borrow' probability mass
for such  `infrequent' words (target words) from those words that are 
in the shortlist and 
semantically close to the target words, as advocated in \cite{khassanov2018unsupervised}.The benefit of using the method in \cite{khassanov2018unsupervised}
is that we don't have to learn a big NLM. However, the downside of 
the method is that word semantic clustering is needed beforehand.
This can be done with word embedding vector. 
For the more details of implementation, one can refer to \cite{khassanov2018unsupervised}.

\section{Experimental setup}
\label{sec:exp}
\subsection{Data}
\label{sssec:data}
We conduct experiments on the SEAME corpus \cite{lyu2010seame} which is a  spontaneous 
conversational
Mandarin-English code-switching speech corpus. 
The duration of utterances that contain code-switching
is about 68\% in our training set.
We define two test sets that contains code-switching speech \cite{GuoXXC18}. Detailed division of SEAME Corpus can be seen in Table \ref{exp:corpus}. One is biased to Mandarin
speech (denoted as {\textit{$dev_{man}$}}), and another
is biased to Southeast Asian accent English ({\textit{$dev_{sge}$}}). 
Each test set contains 10 speakers with balanced genders\footnote{Our KALDI format based test sets are released in the following link: https://github.com/zengzp0912/SEAME-dev-set}. 
In what follows, we report token error rate (TER, Chinese character and English word
respectively) on the test data.

\begin{table}[htbp]
\caption{The detailed division of SEAME Corpus,\textbf{`Man'}, \textbf{`En'} and \textbf{`CS'} mean pure Mandarin, pure English and Mandarin-English code-switch inside utterance.}
\centering
\begin{tabular}{l|l|l|l|l|l}
\hline
\multicolumn{1}{c|}{\multirow{2}{*}{}} & \multicolumn{1}{c|}{\multirow{2}{*}{Speakers}} & \multicolumn{1}{c|}{\multirow{2}{*}{Hours}} & \multicolumn{3}{c}{Duration Ratio}                                          \\ \cline{4-6} 
\multicolumn{1}{c|}{}                  & \multicolumn{1}{c|}{}                          & \multicolumn{1}{c|}{}                       & \multicolumn{1}{c|}{Man} & \multicolumn{1}{c|}{En} & \multicolumn{1}{c}{CS} \\ \hline
train                                   & 134                                            & 101.13                                      &     16\%                     &        16\%                 &          68\%               \\ \hline
{\textit{$dev_{man}$}}                               & 10                                             & 7.49                                        &         14\%                 &        7\%                 &         79\%                \\ \hline
{\textit{$dev_{sge}$}}                                & 10                                             & 3.93                                        &         6\%                 &          41\%               &         53\%                \\ \hline
\end{tabular}
\label{exp:corpus}
\end{table}

\subsection{DNN-HMM baseline system}
\label{sssec:baselineKaldi}
Besides two baselines in Section \ref{sssec:subword}, we also build a DNN-HMM system as another baseline for comparison. We use Kaldi toolkit \cite{povey2011kaldi}
to train a lattice-fee maximum multual information (MMI) based time delay neural network (TDNN) \cite{DBLP:conf/interspeech/PoveyPGGMNWK16}.
The TDNN has 6 hidden layers with 1024 hidden units, and the input features
are 40 dimensional MFCC plus 100 dimensional i-vectors.
The outputs are senones that are langauge independent.
For language modeling, we only use the transcriptions of the training part of the SEAME data to train the 4-gram language model. 

\subsection{E2E ASR system setup}
\label{sssec:e2eSystem}
We use ESPnet toolkit\cite{watanabe2018espnet}\footnote{https://github.com/espnet/espnet} to train our E2E ASR system. The encoder consists of one-layer CNN and six-layers BLSTM with 320 hidden units.  The decoder consists of one-layer LSTM with 320 hidden units.  CTC weight $(1 - \lambda_{1})$ is fixed with 0.2. The attention method used in this work
is a combination of content-based and location-based methods \cite{NIPS2015_5847}. 
To train a BPE subword model, setting character coverage rate for 0.9995 to determine the minimum Mandarin-English mixed symbols, which results in minimum 1806 character symbols. Since we attempt to build 4 BPE subword models with the vocabulary size 1.9k, 2k, 3k and 4k units, all that have dictionary bigger than 1806.


\section{Experimental results and analysis}
\label{sec:exp_results}

\subsection{Results of the E2E CSSR system}
\label{sssec:subsubhead}
Table \ref{exp:e2e_vs_kaldi} reports our TER results 
of the E2E CSSR system with different setups, using 
those from the
Kaldi LF-MMI TDNN CSSR system as a contrast.

\begin{table}[htbp]
\caption{The TER of different E2E CSSR systems as compared to the LF-MMI TDNN ASR counterparts.}
\centering
\begin{tabular}{l|l|l|l|l}
\hline
\multirow{2}{*}{System} & \multirow{2}{*}{\begin{tabular}[c]{@{}l@{}}Data \\ Aug\end{tabular}} & \multirow{2}{*}{Subword} & \multicolumn{2}{l}{TER (\%)} \\ \cline{4-5} 
                        &                                                                               &                          & {\it dev}$_{man}$      & {\it dev}$_{sge}$      \\ \hline
Kaldi-TDNN              & No                                                                            & N.A                      & 23.5          & 32.0          \\ \hline
Kaldi-TDNN-DA           & Yes                                                                           & N.A                      & \textbf{22.1}          & \textbf{30.9}          \\ \hline
\hline 
E2E-CHAR                & No                                                                            & Character               & 34.5          & 46.4          \\ \hline
E2E-DA-CHAR             & Yes                                                                           & Character              & 26.5          & 38.4          \\ \hline
E2E-DA-BPE1.9k            & Yes                                                                           & BPE                      & 26.7          & 36.3          \\ \hline
E2E-DA-BPE2k            & Yes                                                                           & BPE                      & 26.6          & 35.9          \\ \hline
E2E-DA-BPE3k            & Yes                                                                           & BPE                      & 26.4          & 36.1          \\ \hline
E2E-DA-BPE4k            & Yes                                                                           & BPE                      & 26.6          & 36.2          \\ \hline
\end{tabular}
\label{exp:e2e_vs_kaldi}
\end{table}

We have three observations from in Table \ref{exp:e2e_vs_kaldi}.  Firstly, our E2E ASR systems are not as competitive as the LF-MMI TDNN systems in general. 
This suggests that further effort is required to improve
the E2E ASR system at least on the limited training data.
Secondly, data augmentation significantly helps on TER reduction
for our E2E ASR system, suggesting that more data might further
reduce the performance gap between our E2E and LF-MMI TDNN systems.
Thirdly, the BPE subword units are much more
effective than the character units and the 3k BPE produces best results.

One of the differences 
between the CSSR and the monolingual ASR
is that there is cross-lingual substitutions for the CSSR.
Table \ref{exp:error_types} reports various kinds of substitution rates from
different E2E systems.
We note from Table \ref{exp:error_types} that
there are much fewer cross-lingual substitution than same language substitutions, and they are about 10\% on each test sets.
Besides, it can be seen that $S_{E\rightarrow M}$ is higher than $S_{M\rightarrow E}$ by $\sim$5\% which indicates that more English words is substituted by Mandarin characters than the other way. 

\begin{table}[htbp]
\caption{Different substitution (Sub) errors, where \textbf{`M'} and \textbf{`E'} stand for Mandarin and English respectively.}
\centering
\begin{tabular}{c|l|l|l}
\hline
     Sub       &   System         & {\it dev}$_{man}$ (\%) & {\it dev}$_{sge}$ (\%) \\ \hline
\multirow{2}{*}{$S_{E \rightarrow E}$} & E2E-DA-CHAR &  30.8   &  38.5   \\ \cline{2-4} 
                     & E2E-DA-BPE-3k    & 20.9    &  25.3   \\ \hline
\multirow{2}{*}{$S_{E \rightarrow M}$} & E2E-DA-CHAR &  8.4   &  5.0   \\ \cline{2-4} 
                     & E2E-DA-BPE-3k     & 8.0    &  5.2   \\ \hline
\multirow{2}{*}{$S_{M \rightarrow E}$} & E2E-DA-CHAR & 2.7    &  5.2   \\ \cline{2-4} 
                     & E2E-DA-BPE-3k     &  3.0   & 5.1    \\ \hline
\multirow{2}{*}{$S_{M \rightarrow M}$} & E2E-DA-CHAR &  12.9   &  14.7  \\ \cline{2-4} 
                     & E2E-DA-BPE-3k     & 13.0    &  14.7   \\ \hline
\end{tabular}
\label{exp:error_types}
\end{table}

Table \ref{exp:top5errors} shows 
the top 5 examples of cross-lingual substitution on both test sets.
From Table \ref{exp:top5errors}, most of the cross-lingual substitution
are only involved with those non-content or acoustically similar
modal words that are rather short.

\begin{table}[htbp]
\caption{Top 5 cross-lingual substitution examples, where \textbf{`M'} and \textbf{`E'} stand for Mandarin and English respectively.}
\centering
\begin{tabular}{l|l|l|l|l|l}
\hline
\multicolumn{3}{l|}{$S_{M \rightarrow E}$} & \multicolumn{3}{l}{$S_{M \rightarrow E}$} \\ \hline
Ref       & Hyps   & count   & Ref   & Hyps      & count   \\ \hline
ah        & \mman{啊}      & 87      & \textbf{\mman{的}}     & \textbf{the}       & 39      \\ \hline
eh        & \mman{诶}      & 75      & \mman{咯}     & lor       & 36      \\ \hline
er        & \mman{呃}      & 38      & \mman{哦}    & oh        & 21      \\ \hline
\textbf{the}       & \textbf{\mman{的}}      & 37      & \textbf{\mman{有}}     & \textbf{you}       & 18      \\ \hline
oh        & \mman{哦}      & 36      & \mman{诶}     & eh        & 17      \\ \hline
\end{tabular}
\label{exp:top5errors}
\end{table}

\begin{figure*}[htbp]
  \centering
  \includegraphics[width=\linewidth]{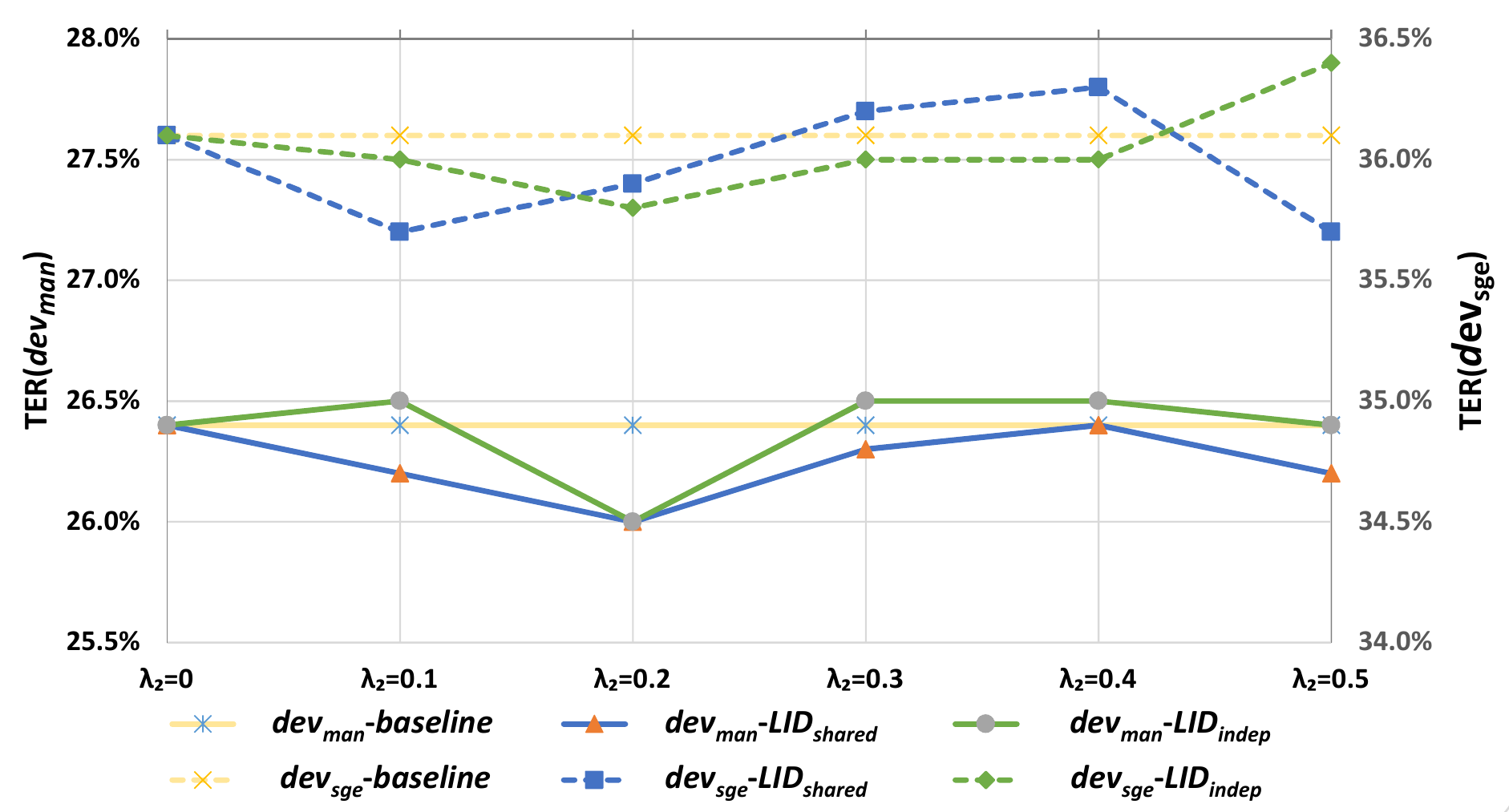}
  \caption{TER results of the MTL with LID for E2E CSSR: TER versus LID weighting factor ($\lambda_2$), {\it dev}$_{man}$ is indicated by left axis, while {\it dev}$_{sge}$ is indicated by right axis.}
  \label{fig:e2e_ter}
\end{figure*}

\subsection{Effect of MTL with LID}
\label{sssec:perforMTL}

Figure \ref{fig:e2e_ter} shows the TER of the two MTL with LID for E2E CSSR methods, {\it LID}$_{shared}$ and
{\it LID}$_{indep}$, at different weighting factor $\lambda_{2}$
in Eq (\ref{eqn:mtl-loss}). The baseline results are the corresponding 
best results from Table \ref{exp:e2e_vs_kaldi}, i.e. E2E-DA-CHAR and E2E-DA-BPE3k.
We observe that the two methods improve the results in most cases, and 
both methods can yield the best results
when $\lambda_{2}$ is around 0.2.  However, there are
no obvious difference between the two methods.

Table \ref{exp:MTL2} reveals
the TER results of the two E2E MTL framework with LID  CSSR methods with $\lambda_{2}=0.2$.
Overall, the proposed methods yield improved results though not significant.
It seems more effective when it is applied to the character based system.

\begin{table}[htbp]
\caption{TER(\%) results of the two E2E MTL framework with LID  CSSR methods ($\lambda_{2} = 0.2$). \textbf{`Man'}, \textbf{`En'} and \textbf{`ALL'} mean English, Mandarin and total TER of the test sets respectively. }
\centering
\begin{tabular}{l@{}|l|l|l|l|l|l}
\hline \multirow{2}{*}{Systems} & \multicolumn{3}{c|}{{\it dev}$_{man}$ (\%)} & \multicolumn{3}{c}{{\it dev}$_{sge}$ (\%)}  \\ \cline{2-7}
 & \multicolumn{1}{c|}{Man} & \multicolumn{1}{c|}{En} & \multicolumn{1}{c|}{ALL} & \multicolumn{1}{c|}{Man} & \multicolumn{1}{c|}{En} & \multicolumn{1}{c}{ALL} \\ \hline
E2E-DA-CHAR  & 21.8   & 39.2   & 26.5   & 28.1   & 44.2   & 38.4   \\ \hline
+ $LID_{shared}$ & 21.8 & 38.7 & 26.3   & 27.7 & 43.6 & 37.9 \\ \hline
+ $LID_{indep}$ & \textbf{21.0} & 38.0 & \textbf{25.6}   & \textbf{27.3} & 42.4 & 37.0 \\ \hline
\hline
E2E-DA-BPE3k  & 22.3   & 37.2   & 26.4   & 28.1   & 40.5   & 36.1   \\ \hline
+ $LID_{shared}$ & 21.9  & \textbf{37.0}  & 26.0  & 27.8  & 40.4  & 35.9  \\ \hline
+ $LID_{indep}$ & 21.8  & 37.3  & 26.0   & 27.7  & \textbf{40.3}  & \textbf{35.8} \\ \hline
\end{tabular}
\label{exp:MTL2}
\end{table}

\subsection{Effect of the NLM vocabulary expansion}
\label{sssec:perforLM}
Table \ref{tab:nlm-exp} reports the TER results of 
the N-best (N=30) rescoring with
the  NLM vocabulary
expansion method, where the NLM is the RNN-LM in practice.
We see from Table \ref{tab:nlm-exp} that the proposed NLM
vocabulary expansion method achieved consistent improved TER results 
over the best results shown in the last row of Table \ref{exp:MTL2}.
Finally, experiment results showed that applying the proposed approaches significantly reduces the TER of two dev sets from 34.5\% and 46.5\% to 25.0\% and 34.5\% respectively, which is close to the results of strong LF-MMI TDNN system.

\begin{table}[htbp]
\caption{TER of the NLM vocabulary expansion.}
\centering
\begin{tabular}{l|l|l}
\hline
Method      & {\it dev}$_{man}$ (\%)    & {\it dev}$_{sge}$ (\%)  \\ \hline
NoLM            & 26.0    & 35.8    \\ \hline
5-gram KN       &  25.9   &  35.6   \\ \hline
RNN-LM           &  25.1   &   34.6  \\ \hline
 \hline 
Proposed NLM     &  \textbf{25.0}  &  \textbf{34.5}   \\ \hline 
\end{tabular}
\label{tab:nlm-exp}
\end{table}

\section{Conclusion and future work}
\label{sec:conclusion}
In this paper we proposed several approaches to improve E2E based 
Mandarin-English code-switching speech recognition. 
These approaches include data augmentation, byte-pair encoding subword units for English language, language identification based multitask learning,
as well as the vocabulary expansion for neural language models to rescore the N-best results. 

In the future, we plan to study leveraging external monolingual data to improve its performance. we also plan to incorporating language model into the existing model to improve its performance. 

\section{Acknowledgment}
\label{sec:page}
This work is supported by the project of Alibaba-NTU Singapore Joint Research Institute.

\bibliographystyle{IEEEtran}
\newpage
\bibliography{mybib}

\end{document}